\documentclass[10pt,twocolumn,letterpaper]{article}

\usepackage{3dv}
\usepackage{times}
\usepackage{epsfig}
\usepackage{graphicx}
\usepackage{graphicx}
\usepackage{amsmath,amssymb} 
\usepackage{color}
\usepackage{stfloats}
\usepackage{tablefootnote}
\usepackage{amsmath}
\usepackage{amssymb}
\usepackage{algorithm}
\usepackage{algorithmic}
\usepackage{booktabs}
\usepackage{array}
\usepackage[pagebackref=true,breaklinks=true,letterpaper=true,colorlinks,bookmarks=false]{hyperref}

\threedvfinalcopy 

\ifthreedvfinal\fi
\begin{document}

\title{Generative Adversarial Frontal View to Bird View Synthesis}

\author{Xinge Zhu$^{\dag}$~~~~~Zhichao Yin$^{\S}$~~~~~Jianping Shi$^{\S}$~~~~~ Hongsheng Li$^{\dag}$~~~~~Dahua Lin$^{\dag}$~~~~~\\ $^{\dag}$CUHK-SenseTime Joint Lab, CUHK\\$^{\S}$SenseTime Research \\ {\tt\small {\{zhuxinge, yinzhichao, shijianping\}@sensetime.com}} \\ {\tt\small hsli@ee.cuhk.edu.hk, dhlin@ie.cuhk.edu.hk}}


\maketitle
\thispagestyle{empty}
\begin{abstract}
\label{sec:abs}
Environment perception is an important task with great practical value and bird view is an essential part for creating panoramas of surrounding environment. 
Due to the large gap and severe deformation between the frontal view and bird view, generating a bird view image from a single frontal view is challenging. To tackle this problem, we propose the BridgeGAN, i.e., a novel generative model for bird view synthesis. First, an intermediate view, i.e., homography view, is introduced to bridge the large gap. Next, conditioned on the three views (frontal view, homography view and bird view) in our task, a multi-GAN based model is proposed to learn the challenging cross-view translation. 
Extensive experiments conducted on a synthetic dataset have demonstrated that the images generated by our model are much better than those generated by existing methods, with more consistent global appearance and sharper details. 
Ablation studies and discussions show its reliability and robustness in some challenging cases. Codes are available at \url{https://github.com/WERush/BridgeGAN}

\end{abstract}

\section{Introduction}
\label{sec:intro}



View synthesis is a long-standing problem in computer vision~\cite{furukawa2015multi,ji2017deep,cross,yan2016perspective,predict}, which facilitates many applications including surrounding perception and virtual reality. In modern autonomous driving solution, the limited viewpoint of on-car cameras restricts the system from reliably understanding the environment, acquiring accurate global view for better policy making and path planning. 
Due to the large view gap (90 degrees in our task) and severe deformation, generating the bird view image from the front view is not even easy for human being. 
In this paper, we would like to push the envelop of synthesis between two drastically different views, although the challenging nature of this problem leaves room for further improvements.
To our best knowledge, it is the {first} attempt to generate the bird view based on single frontal view image, which serves better perceptual understanding and sparks future researchers to explore information from multiple views for perception.



\begin{figure}[t]
\begin{center}
   \includegraphics[width=0.8\linewidth]{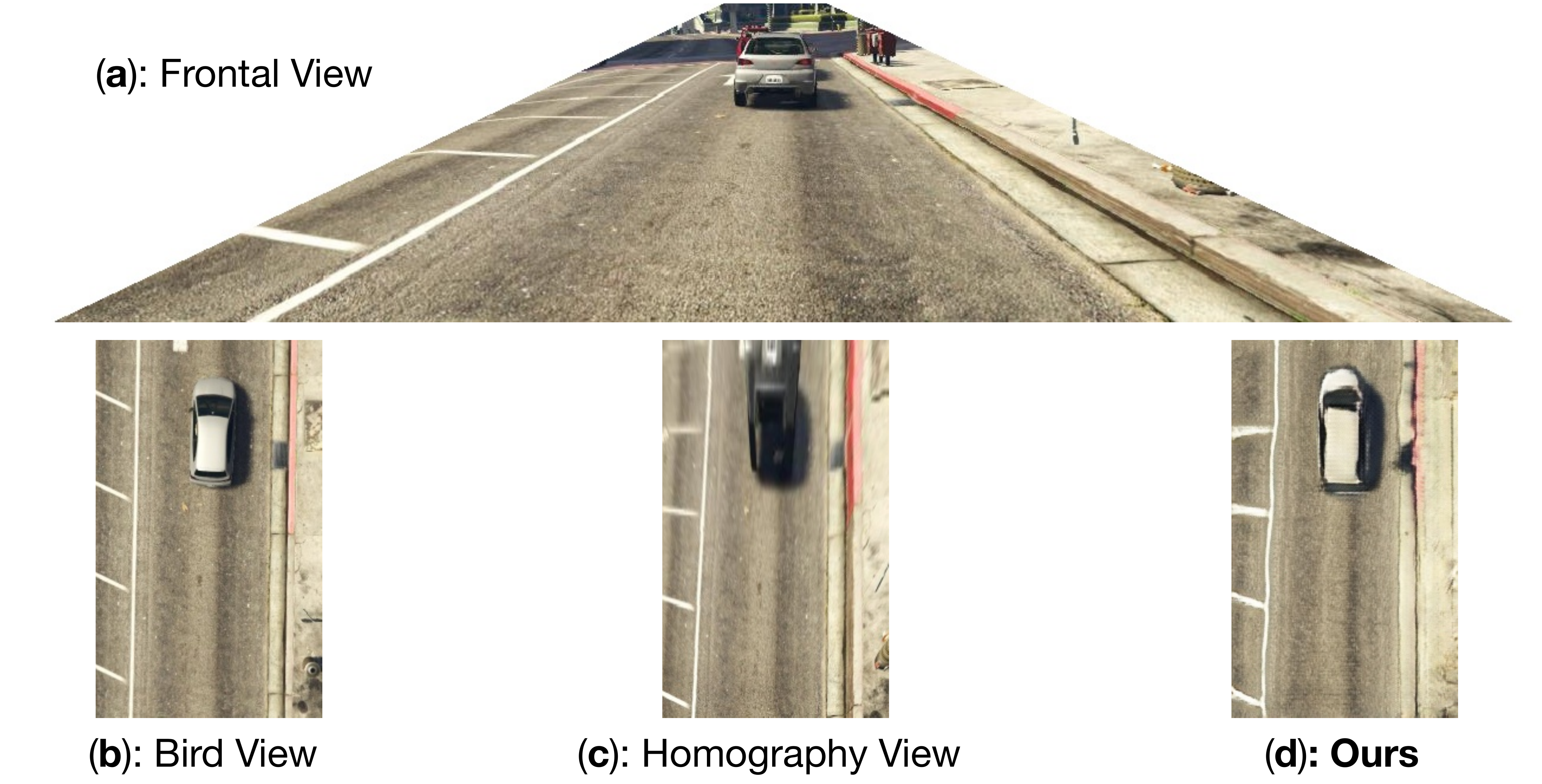}
\end{center}
\vspace{-3ex}
\caption{(\textbf{a}): The frontal view image. (\textbf{b}): The ground truth bird view image. (\textbf{c}): The intermediate homography view image. {(\textbf{d})}: The generated bird view image by our model.}
\label{fig:intro}
\vspace{-4ex}
\end{figure}

With the 3D structure of the scene, bird view generation can be easily achieved by changing the view point and projection. However, when the only input is a frontal view image, it will be substantially more difficult. 
The same object has different appearances and sizes in the images of bird view and frontal view. Meanwhile, the semantic representation and basic color should be consistent between two views. Imagining a car in front of you, after transforming to bird view, it should be the same car but with completely different appearance and size. An example is shown in Fig.~\ref{fig:intro} (a) and (b). The large gap and severe deformation make it a challenging task and it is far from being solved.


Traditional methods for bird view generation are generally built on multi-sensor systems
\cite{nielsen2005surround,sung2012development,zhang2014surround}. Recently, by regarding the view transformation as a view synthesis task, many 3D view synthesis methods obtained promising results by modeling the underlying 3D geometry~\cite{furukawa2015multi,yan2016perspective}. Image-based rendering models \cite{deepstereo,zhou2016view}, on the other hand, generate new views by re-using the pixels from source images. However, these methods can only transform already visible content, e.g., they cannot render the top view of a car from the input frontal view or side-view images. 


Given the large gap between the frontal view and the bird view images, they can be naturally regarded as two different domains. 
With the vigorous study of generative adversarial network~\cite{goodfellow2014generative}, many powerful cross-domain image translation systems~\cite{discoGan,yi2017dualgan,zhang2016colorful,cycleGan,Zhu2018PenalizingTP} have been proposed, which can generate images with plausible appearance. 
The representative work pix2pix~\cite{pix2pix} utilizes a conditional adversarial network, converting an image from one representation of a given scene to another, e.g. semantic labels to images, edge-map to photograph. However, these models could only perform translations for the aligned images in color or texture level, e.g. from zebra to horse or from grey-scale to color. 
Translating images across two domains with a large gap in between (e.g.~the images captured from different viewpoints) remains a challenging task even with the latest GAN-based techniques.


In this paper, we propose the \textbf{BridgeGAN} model, a novel bird view generation model from single frontal view images. 
To bridge the large gap between the two views, we incorporate the homography view as the intermediate view, with a homography matrix \cite{homography} to perform the perspective mapping, as shown in Fig.~\ref{fig:intro} (c). It serves as a bridge to connect two views. 
Hence, this is where the `Bridge' comes from. 
The homography view serves to decrease the gap between frontal view and bird view, but it produces undesired distortions. 
Conditioned on the three views in our task, a multi-GAN based model is proposed to learn the cross-view translation. 
We extend the cycle-consistency loss \cite{cycleGan} to a dual cycle-consistency loss for matching the three views and constraining the cross-domain translation to be a one-to-one correspondence. Furthermore, a cross-view feature consistency loss is designed to make all three views have a shared feature representation in low (e.g. color) and high (e.g. content) level. The final result is shown in Fig.~\ref{fig:intro}~(d).
Experimental results demonstrate that, by generating images consistent in terms of global structure and details, our method results in significantly better performance compared to the baselines.
Ablation studies verify the effect of each components, and discussions show the reliability of our model in the case of challenging scenario. 

The main contributions are summarized as follows:
\begin{itemize}
\item[(1)] To our best knowledge, we are the {first} to address the novel problem of generating bird view image based on a single frontal view image, which enables better perceptual understanding and multiple views perception.
\item[(2)] We propose the BridgeGAN, i.e., a novel generative model for bird view synthesis, in which the homography view is first introduced to bridge the gap and then a multi-GAN based model is proposed to perform cross-view transformation. 
\item[(3)] Extensive experiments demonstrate that the proposed model generates significantly better results compared with baselines, which is able to preserve the global appearance and the details of objects. More discussions also verify its reliability and robustness.
\end{itemize}

\section{Related Work}
\label{sec:related}
\noindent \textbf{Bird-Eye View}~~~There are few works in literature that aim to tackle the problem of perspective transformation. Most of these methods are geometry based. More specifically, Lin~\emph{et al.}~\cite{lin2012vision} introduced a fitting parameters searching algorithm to estimate a perspective matrix for image coordinate transformation. Similarly, in \cite{tseng2013image}, an inverse perspective matrix was used to perform the view transformation. However, the largest problem of such kind of methods is the distortion, especially the region in a distance. Another group of methods are vision based. \cite{liu2014photometric,zhang2014surround} achieved a bird-eye view by stitching images from a four to six fish-eye lens cameras system. Sung~\emph{et al.} \cite{sung2012development} proposed a camera parameter optimization algorithm to establish surround view from multi-camera images. However, in most cases, a multi-camera system is not available in the vehicle and the common source is the frontal view image. Moreover, such methods cannot create new view invisible from existing cameras.

\noindent \textbf{View Synthesizing}~~~A large body of view synthesizing works are geometry based. With a huge amount of multi-view images, 3D stereo algorithms \cite{furukawa2015multi} are applicable to reconstruct the 3D scene and then be utilized to synthesize novel views. Ji \emph{et al.} \cite{ji2017deep} proposed to synthesize middle view images by using two rectified view images. Yan \emph{et al.} \cite{yan2016perspective} proposed a perspective transformer network to learn the projection transformation after reconstructing the 3D volume of the object.
However, most of these methods are trained with 3D supervision and all view pairs can be generated via a graphics engine. In our setting, the training data is limited in both views and numbers and the 3D supervision is also unavailable.

\noindent \textbf{Synthesis between Remote Sensing Image and Ground-level Image}~~~
Existing works~\cite{predict,cross} explored to predict the semantic segmentation of ground level image from its remote sensing image, and then apply the semantic layout to synthesize the ground image. However, the remote sensing image and ground-level image have the much higher viewpoint and lack of textures, which are significantly different from our on-car camera image and bird view image. Besides, these methods also require the segmentation mask to supply the synthesis, which is not feasible in our setting. Furthermore, due to the little or no overlap between remote sensing image and ground image, this view synthesis desires the model to imagine a large unseen region while our setting has more proper inference with more overlap. 


\vspace{1ex}
\noindent \textbf{Image Generation}~~~Recently, image generation has been a heated topic with the emerge of Generative Adversarial Networks (GANs) \cite{goodfellow2014generative}. A GAN consists of two modules, a generator $G$ and a discriminator $D$. 
These two parts $G$ and $D$ play a minimax game. $G$ is trained to generate images to confuse the $D$, and $D$ is trained to distinguish between real and fake samples.

Various methods were developed to generate images based on GANs. Conditional generative adversarial nets (CGANs) \cite{mirza2014conditional} used the labels as the conditional information to both generator and discriminator and generated digit images of specified label. Pix2pix model \cite{pix2pix} was  built on the CGANs to learn the mapping with full supervision. Similar ideas have been applied to numerous tasks such as generating images from sketches or from text \cite{reed2016generative,sangkloy2016scribbler,zhang2016stackgan}.
Many researchers also tried to attempt multiple GANs for learning a mapping from input domain to target domain. CoupledGANs \cite{liu2016coupled} used a weight-sharing strategy to learn a common representation across domains. CycleGAN \cite{cycleGan} and DiscoGAN \cite{discoGan} introduced an inverse mapping and a cycle consistency loss to constrain the mapping between two domains. DualGAN \cite{yi2017dualgan} and UNIT~\cite{unit} also employed another GAN to learn to invert the image translation task. Our model is also multi-GAN based. But unlike these image translation approaches with two GANs, our model applies three GANs with two consistency constraints to strengthen the generation capability due to the large gap between the frontal view and the bird view image.

\section{Methodology}
\label{sec:method}

\begin{figure*}[ht]
\begin{center}
   \includegraphics[width=1.0\linewidth]{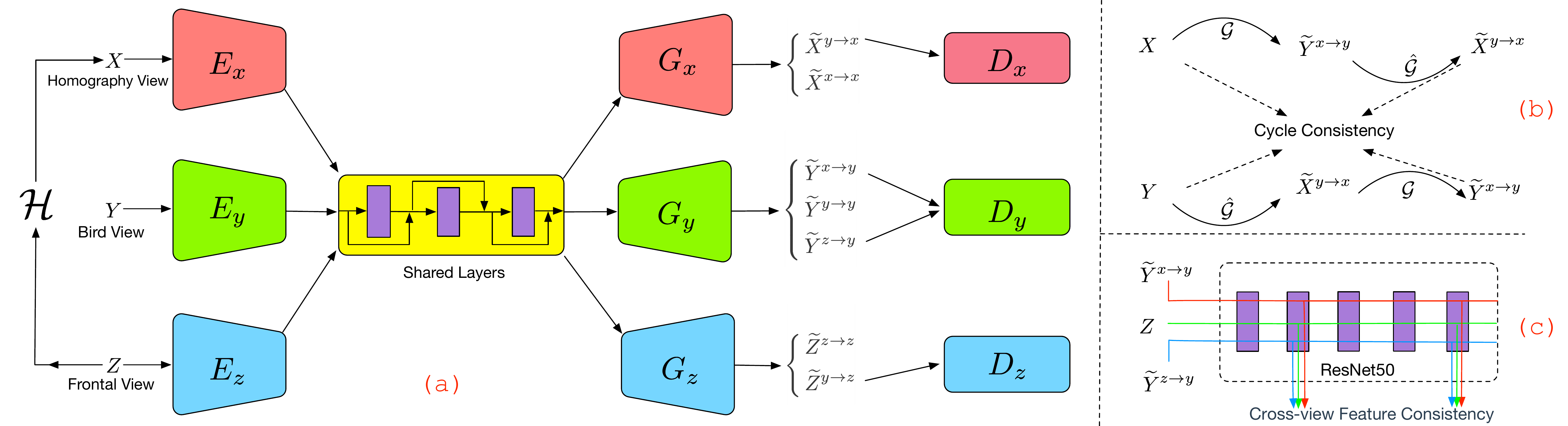}
\end{center}
\vspace{-2ex}
   \caption{\textcolor{red}{(a)} Our model contains three GANs representing three domains, i.e. homography view($X$), Bird view($Y$) and Frontal view($Z$). Each of them consists of three modules: Encoder($E$), Generator($G$) and Discriminator($D$). Since all three domains have a shared semantic meaning, we employ shared layers for learning unified intermediate representations. $G$s learn the cross-domain translations and reconstruction mappings based on the intermediate representation. $D$s are adversarial discriminators for the respective domains, distinguishing between real and fake samples. \textcolor{red}{(b)} Single cycle-consistency loss includes two components, i.e. forward cycle: $X\xrightarrow[]{\mathcal{G}} \widetilde{Y}^{x\to y} \xrightarrow[]{\hat{\mathcal{G}}} \widetilde{X}^{y\to x} \approx X$ and backward cycle: $Y\xrightarrow[]{\hat{\mathcal{G}}} \widetilde{X}^{y\to x} \xrightarrow[]{{\mathcal{G}}} \widetilde{Y}^{x\to y} \approx Y$. \textcolor{red}{(c)} To enforce the consistent cross-view features, we introduce the loss network (ResNet-50) for regularizing the inconsistent problem.}
\label{fig:pipeline}
\vspace{-1ex}
\end{figure*}


Bird view synthesis can be viewed as a cross-domain image generation task, in which the source domain is frontal view and target domain is bird view. Unlike these traditional cross-domain image translation tasks that are performed between two aligned images \cite{discoGan,cycleGan}, such as from zebras to horses and photos to paintings, bird view synthesis is more challenging because the large gap and severe deformation exist between frontal view and bird view. To tackle this challenge, we introduce an intermediate view, homography view, to bridge the gap in our task, then a multi-GAN based model is proposed to realize the cross-domain translation. 



\subsection{Framework Overview}
Our full BridgeGAN is illustrated in Fig.~\ref{fig:pipeline} (a).
There are three domains: $X$ for homography view, $Y$ for bird view and $Z$ for frontal view. The BridgeGAN learns by synthesizing one view from another via GAN, enabling the network to learn the intermediate representation shared between different views and the reconstruction ability upon the intermediate representation.

More specifically, the BridgeGAN model consists of three GANs for the three domain representations and translations as a multi-GAN system. Each GAN contains an Encoder ($E$) to transform image to an intermediate representation, and a Generator ($G$) to transform the intermediate representation to a new image. In addition, there exists a discriminator ($D$) to distinguish between generated images and real images, which drives the generative image towards real one. 

During training, the bird view domain is chosen as a pivot and two cross-domain translations exist in the pipeline, i.e. between homography view ($X$) and bird view ($Y$), and between frontal view ($Z$) and bird view ($Y$). 
We introduce shared layers to enforce the GANs to share some higher-layer parameters for a consistent intermediate representation between views. 
After training, the proposed framework generates bird view images by two steps: first doing a homography estimation from $Z$ to $X$ and then performing cross-domain translation from homography view ($X$) to bird view ($Y$).

\begin{table}
\small
\setlength{\tabcolsep}{0.8pt}
\begin{center}
\begin{tabular*}{1.0\linewidth}{c|ccccc}
\toprule
Mappings & $I_x$ & $\mathcal{G}$ & $\mathcal{F}$ & $\mathcal{\hat{G}}$ & $\hat{\mathcal{F}}$\\
\midrule
Subnetworks & $\{E_x,G_x\}$&$\{E_x,G_y\}$&$\{E_z,G_y\}$&$\{E_y,G_x\}$&$\{E_y,G_z\}$\\
\bottomrule
\end{tabular*}
\end{center}
\caption{The relation between mappings and subnetworks in our model. For the mapping $\mathcal{G}:X\to Y$, it is composed of two subnetworks, \ie Encoder $E_x$ and Generator $G_y$. In addition, 
$I_y$ and $I_z$ have the same formation as $I_x$.}
\label{tab:mappings}
\end{table}

\subsection{Multi-GAN Learning}
We design the multi-GAN system to learn the cross-domain translations for bird view generation from both frontal view and homography view.


 
\vspace{9pt}
\noindent \textbf{View Transformation.}~~~Our model builds upon seven view mappings between all three domains in our multi-GAN system, including cross-domain mappings $\mathcal{G}: X\to Y$; $\mathcal{F}: Z\to Y$, their inverse mappings $\hat{\mathcal{G}}: Y\to X$; $\hat{\mathcal{F}}: Y \to Z$ and three identity mappings $I_x, I_y, I_z$. The cross domain mapping captures the view transformation ability whereas the identity mapping ensures the reconstruction consistency. Specified, the homography view and frontal view are complementary parts to realize the bird view translation, where the homography view decreases the gap from frontal view to bird view and the frontal view provides more global context that homography view lacks of in turn. We do not include mapping between frontal view $Z$ and homography view $X$ since it is a fixed perspective transformation. Each view transformation is represented by a GAN. Table \ref{tab:mappings} summarizes the mappings and corresponding implementations in our model.

 
\vspace{9pt}
\noindent \textbf{Full Objective.}~~~Our objective contains three terms: \emph{adversarial loss} $\mathcal{L}_{\text{GAN}}$ for matching the distribution of generated image to the distribution of target image, \emph{dual cycle-consistency loss} $\mathcal{L}_{\text{cyc}}$ for constraining the cross-domain mappings to be a one-to-one correspondence and to be well covered on bi-directions (bijective mappings), and \emph{cross-view feature consistency loss} $\mathcal{L}_{\text{cfc}}$ for encouraging generated bird view image and frontal view image to keep the feature representations consistent in low and high level, such as color and content. Our full objective is:
\begin{equation}
\begin{split}
\mathcal{L} = \mathcal{L}_{\text{GAN}} + \mathcal{L}_{\text{cyc}} + \mathcal{L}_{\text{cfc}}.
\end{split}
\end{equation}


\noindent \textbf{Adversarial Loss.}~~~
Inheriting from GAN \cite{goodfellow2014generative}, we apply the adversarial losses to these cross-domain mappings, i.e. $\mathcal{G}$, $\mathcal{F}$, $\hat{\mathcal{G}}$ and $\hat{\mathcal{F}}$. This objective enforces our model to learn mapping from its input domain to target domain. 

For the mapping $\mathcal{G}: X\to Y$, we express the objective as: 
\begin{equation}
\begin{split}
\mathcal{L}_{\text{GAN}}(\mathcal{G}, D_y, X,Y) &= \mathbb{E}_{y\sim p_{\text{data}}(y)}[\log(D_y(y)]\\ &+ \mathbb{E}_{x\sim p_{\text{data}}(x)}[\log(1-D_y(\mathcal{G}(x)))],
\end{split}
\end{equation} 
where $\mathcal{G}$ aims to generate the image that looks similar to image from target domain $Y$, while $D_y$ tries to identify the real and fake samples. 
Similarly, we introduce this objective for other cross-domain mappings, i.e. $\mathcal{F}$, $\hat{\mathcal{G}}$ and $\hat{\mathcal{F}}$. Hence, we get the objectives, $\mathcal{L}_{\text{GAN}}(\mathcal{F}, D_y, Z,Y)$, $\mathcal{L}_{\text{GAN}}(\hat{\mathcal{G}}, D_x, Y,X)$ and $\mathcal{L}_{\text{GAN}}(\hat{\mathcal{F}}, D_z, Y,Z)$, respectively. The total adversarial loss is the sums of cross-domain mappings, which is expressed as:
\begin{equation}
\begin{split}
\mathcal{L}_{\text{GAN}} &= (\mathcal{L}_{\text{GAN}}(\mathcal{G}, D_y, X,Y) + \mathcal{L}_{\text{GAN}}(\hat{\mathcal{G}}, D_x, Y,X)) \\ &+ (\mathcal{L}_{\text{GAN}}(\mathcal{F}, D_y, Z,Y) + \mathcal{L}_{\text{GAN}}(\hat{\mathcal{F}}, D_z, Y,Z)).
\end{split}
\end{equation}
Furthermore, these generators are tasked to not only fool the discriminators but also to be near the ground truth image in the pixel level. For pixel-level loss of these generators, we use L1 distance rather L2 as L1 encourages less blurring. 


\vspace{9pt}
\noindent \textbf{Dual Cycle-consistency Loss.}~~~ 
To reduce the space of the possible mapping functions and enforce the mappings between two domains to be a one-to-one bijective mapping, a cycle-consistency loss \cite{discoGan,yi2017dualgan,cycleGan} was introduced. We extend the cycle-consistency loss to a dual cycle-consistency loss for matching the three domains in our task. Each cycle consistency serves a cross-domain mapping and its inverse mapping, which measures how well the origin input is reconstructed after a sequence of two generations. 

The first cycle consistency is for the mapping $\mathcal{G}:X\to Y$ and its inverse mapping $\hat{\mathcal{G}}: Y\to X$. As illustrated in Fig.~\ref{fig:pipeline} (b), we depict the pipeline of forward cycle consistency and backward cycle consistency, i.e. $X\xrightarrow[]{\mathcal{G}} \widetilde{Y}^{x\to y} \xrightarrow[]{\hat{\mathcal{G}}} \widetilde{X}^{y\to x} \approx X$ and $Y\xrightarrow[]{\hat{\mathcal{G}}} \widetilde{X}^{y\to x} \xrightarrow[]{{\mathcal{G}}} \widetilde{Y}^{x\to y} \approx Y$. Thus the first cycle consistency objective can be given by:
\begin{equation}
\begin{split}
\mathcal{L}_{\text{cyc}}(\mathcal{G}, \hat{\mathcal{G}}) &= \mathbb{E}_{x\sim p_{\text{data}}(x)}[\parallel \hat{\mathcal{G}}(\mathcal{G}(x))-x\parallel_1 ]\\ &+ \mathbb{E}_{y\sim p_{\text{data}}(y)}[\parallel\mathcal{G}(\hat{\mathcal{G}}(y))-y\parallel_1],
\end{split}
\end{equation} 
where the L1 norm is applied to measure the distance between $\hat{\mathcal{G}}(\mathcal{G}(x))$ and $x$. 

Similarly, for another cross-domain mapping $\mathcal{F}:Z\to Y$ and its inverse mapping $\hat{\mathcal{F}}$, the second cycle consistency objective is:
\begin{equation}
\begin{split}
\mathcal{L}_{\text{cyc}}(\mathcal{F}, \hat{\mathcal{F}}) &= \mathbb{E}_{z\sim p_{\text{data}}(z)}[\parallel \hat{\mathcal{F}}(\mathcal{F}(z))-z\parallel_1 ]\\ &+ \mathbb{E}_{y\sim p_{\text{data}}(y)}[\parallel\mathcal{F}(\hat{\mathcal{F}}(y))-y\parallel_1].
\end{split}
\end{equation} 
Finally, we express the dual cycle consistency objective function as:
\begin{equation}
\begin{split}
\mathcal{L}_{\text{cyc}} = \lambda_1(\mathcal{L}_{\text{cyc}}(\mathcal{G}, \hat{\mathcal{G}}) + \mathcal{L}_{\text{cyc}}(\mathcal{F}, \hat{\mathcal{F}})).
\end{split}
\end{equation}


\vspace{9pt}
\noindent \textbf{Cross-view Feature Consistency Loss.}~~~Considering two images from frontal view and bird view, the semantic contents and colors are consistent among them. We postulate there exists a cross-view feature consistency among three different views in our task.  However, the adversarial loss could make the generated image look realistic but its contents may not be closely relevant to the source, and the dual cycle-consistency loss enforces a one-to-one constraint on cross-domain mappings rather than a joint consistency on low- and high-level features among three domains. Therefore, we design another consistency loss to enforce the constraint of consistent cross-view feature representations among all three views.  

Since the generated bird view image is not aligned with frontal view, direct pixel-wise loss (L1, L2, etc.) is not suitable. 
We introduce a loss network $\phi$ pretrained for image classification to satisfy this constraint. Instead of minimizing the pixel-level distance, 
we encourage the generated bird view image and real frontal view image to have similar low- and high-level feature representations extracted from loss network $\phi$. 
Feature maps extracted from early layer model the low-level features and feature maps from higher layer serve as high-level representations.
The loss function is the normalized L2 distance:
\begin{equation}
\begin{split}
\mathcal{L}_{\phi}(\widetilde{Y}, Z) &= \frac{1}{C_jH_jW_j}\parallel \phi_j(\widetilde{Y})-\phi_j(Z)\parallel_2^2\\ &+ \frac{1}{C_kH_kW_k}\parallel \phi_k(\widetilde{Y})-\phi_k(Z)\parallel_2^2,
\end{split}
\end{equation}
where $\widetilde{Y}$ and $Z$ are generated bird view and frontal view image, respectively. $j$ and $k$ denote the $j$th and $k$th layer in the loss network. $C$, $H$ and $W$ are the shape of feature maps. The cross-view feature consistency can be given by: 
\begin{equation}
\begin{split}
\mathcal{L}_{\text{cfc}}= \mathcal{L}_{\text{cfc}}(\mathcal{G},\mathcal{F}, Z, \phi) &=  \mathbb{E}_{x\sim p_{\text{data}}(x)}\mathcal{L}_{\phi}(\mathcal{G}(x), Z)\\ &+ \mathbb{E}_{z\sim p_{\text{data}}(z)}\mathcal{L}_{\phi}(\mathcal{F}(z), Z),
\end{split}
\end{equation} 
where $\mathcal{G}(x)$ and $\mathcal{F}(z)$ are the generated bird view images from homography view and frontal view, respectively.
As shown in Fig.~\ref{fig:pipeline} (c),  in our experiments, $\phi$ is the resnet-50 \cite{resnet} pretrained on the ImageNet \cite{imagenet} dataset. We utilize the feature maps from $res2c$ layer as the low-level feature and $res5c$ as the high-level representation.

\subsection{Iterative Optimization}

For the training of GANs, it can be viewed as playing a minimax game until finding a saddle point. In the proposed framework, these encoders and generators are in one team versus another team composed of adversarial discriminators. Our full optimization objective is:
\begin{equation}
\begin{split}
\min \limits_{E_x,E_y,E_z,G_x,G_y,G_z;} \max \limits_{D_x,D_y,D_z} \mathcal{L}_{\text{GAN}} + \mathcal{L}_{\text{cyc}} + \mathcal{L}_{\text{cfc}}.
\end{split}
\end{equation}
We apply an alternative optimization approach, which iteratively updates the network blocks in the following order: (1) Update discriminators: the network is first optimized by maximizing the discriminators' accuracy with encoders and generators fixed. (2) Update encoders and generators by minimizing the loss with discriminators fixed.



\subsection{Implementation Details}
\begin{figure}[t]
\begin{center}
   \includegraphics[width=1.0\linewidth]{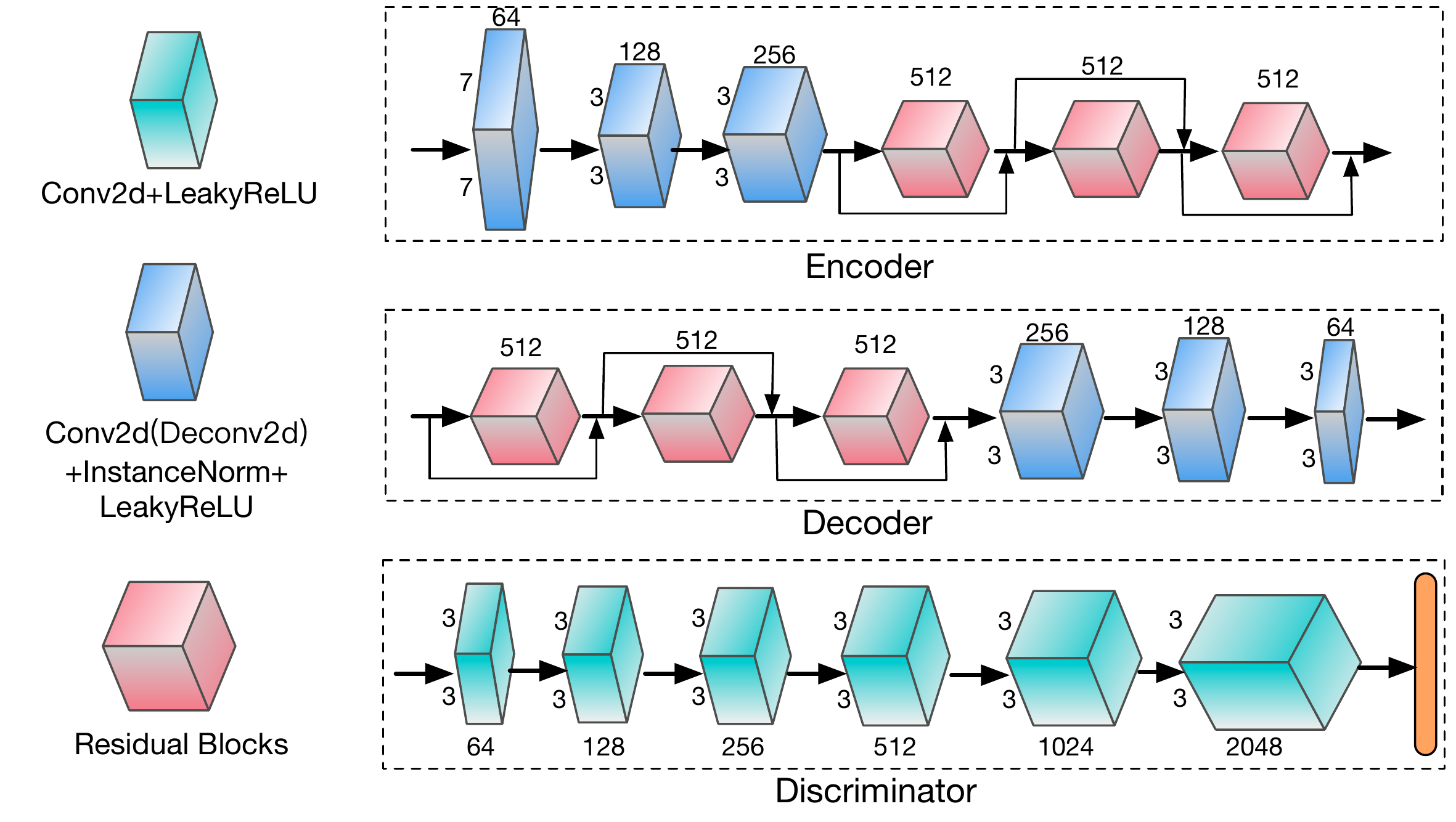}
\end{center}
\vspace{-3ex}
\caption{Detailed architectures of subnetworks in our proposed framework, including encoder, decoder and discriminator. Basic components are shown in the left column. Note that the convolutional layers would be replaced by deconvolutional layers \cite{FCN} in the decoder and the shape of each layer is labelled.}
\label{fig:subnet}
\vspace{-3ex}
\end{figure}

As illustrated in Fig.~\ref{fig:subnet}, we elaborate the structure of encoder, decoder and discriminator in our model. There are three basic components shown in the left column.   
Inheriting from \cite{improved}, we utilize the LeakyReLU and instance normalization \cite{instanceNorm} to improve training stability. The encoder and decoder are both composed of six blocks and detailed architectures are shown in the figure. For the residual block, we use the same architecture as resnet \cite{resnet}.

\vspace{3pt}
\noindent \textbf{Shared Layers.}~~~As shown in Fig.~\ref{fig:pipeline} (a), shared layers
 are composed of three stacked residual blocks \cite{resnet} between encoder and decoder networks. The features from three domains are handled by shared layers for enforcing a unified intermediate representation.

\vspace{3pt}
\noindent \textbf{Homography Estimation.}~~~We follow the traditional homography estimation pipeline \cite{homography} which is composed of two stages: corner estimation and robust homography estimation. For the stage \uppercase\expandafter{\romannumeral1}, we utilize the ORB \cite{ORB} features as the descriptor. For the stage \uppercase\expandafter{\romannumeral2}, a parameter searching algorithm RANSAC \cite{ransac} is used as the estimator. Then a direct linear transform (DLT) algorithm \cite{DLT} is applied to get the homography matrix $\mathcal{H}$. We can obtain the homography view by using the equation $X = \mathcal{H}Z$. 
More details are introduced in Section \ref{sec:id}.

\section{Experiments}
\label{sec:exper}
\subsection{Dataset and Evaluation Metrics}
\noindent \textbf{Dataset.}~~~In \cite{GTAV}, they developed a framework to collect data from Grand Theft Auto V video game, in which the game camera automatically toggles between frontal and bird view at each time step. In this way, they gathered information about the road scene from both views. We download the dataset from their official website\footnote{http://imagelab.ing.unimore.it/scene-awareness}. 
As shown in the Fig.~\ref{fig:intro} (a), we remain the region right in front of the vehicle, which is the common part between the frontal view and the bird view. 
After data processing, we obtain a training set with 40,000 pairs of images and a testing set with 4,000 pairs.

\vspace{1ex}
\noindent \textbf{Evaluation Metrics.}~~~
We use two traditional metrics PSNR, SSIM~\cite{SSIM}, and a neural network based metric LPIPS~\cite{LPIPS}. PSNR relies on low-level differences. SSIM mainly reflects the perceived change in the structural information. LPIPS uses pretrained deep models to evaluate the similarity, which highly agrees well with humans. Specifically, we use the pretrained AlexNet in the LPIPS\footnote{https://github.com/richzhang/PerceptualSimilarity} metrics. 


\subsection{Training Setup}
\label{sec:id}

For the homography estimation, we utilize multi-scale ORB features as the descriptor and select the top 30 scoring matches as the input to the RANSAC estimator. After obtaining the homography matrix, we perform the homography estimation based on the cropped frontal view image to get the homography view image. All three view images are resized to 320x192 as the inputs.

We apply the PatchGAN \cite{patchgan} to the discriminators, in which the discriminator tries to classify whether overlapping image patches are real or fake. Similar to EBGAN \cite{ebgan}, we add the gaussian noise to the shared layers and generator. During training, Adam \cite{adam} optimization is applied with $\beta_1$=0.5 and $\beta_2$=0.999. We train the model on a single Titan X GPU with learning rate=0.0001. Each mini-batch contains three frontal view images, three homography view images and three bird view images. For the weighted factor $\lambda_1$, we apply $\lambda_1 = 10$, which is chosen by using a cross-validation method.  

\subsection{Experimental Results}

\noindent \textbf{Baseline methods}~~~
\noindent \emph{\textbf{Pix2pix}} \cite{pix2pix}. This method proposes a conditional GAN for image translation. Without hand-engineering loss functions, the conditional adversarial networks could also achieve reasonable results. 

\noindent \emph{\textbf{CycleGAN}}~\cite{cycleGan}. Like our method, CycleGAN is also a multi-GAN model. It consists of two GANs representing two domains and a cycle-consistency loss is used to regularize the cross-domain mapping.

\noindent \emph{\textbf{DiscoGAN}}~ \cite{discoGan}. DiscoGAN is another multi-GAN based model which is proposed to discover the cross-domain relations.

\noindent \emph{\textbf{CoGAN}}~\cite{liu2016coupled}.
This method consists of two GANs, and a strategy of weights sharing is introduced. Unlike our method sharing the representation in the higher layer, CoGAN shares the weights on its first few layers.

\noindent \emph{\textbf{Homo}}
Since bird view generation is a perspective transformation in geometry, the homography estimation is also a viable solution. We term it as Homo and regard it as a baseline method.

\begin{figure*}[t]
\begin{center}
   \includegraphics[width=1.0\linewidth]{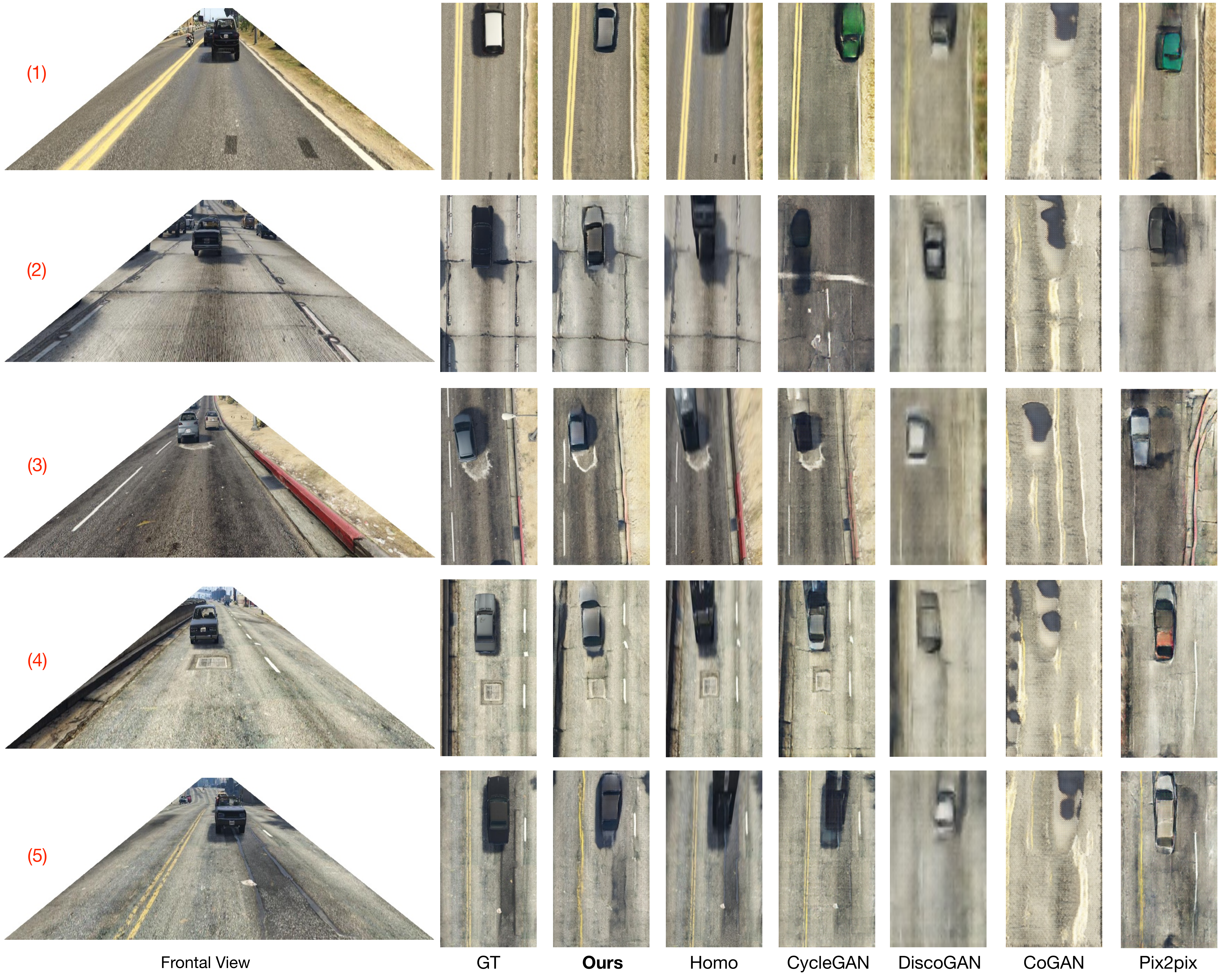}
\end{center}
\vspace{-3ex}
\caption{Example results by our proposed method and baselines. Best viewed in color.}
\label{fig:exper}
\vspace{-2ex}
\end{figure*}

\noindent \textbf{Comparison against baselines} ~~~ 
Since pairs of images are provided in the dataset, we train all baseline methods in a \textbf{supervised} manner for a fairness, which means that during the training of baselines, \textbf{paired images (homography view and corresponding bird-view images) are fed into those baselines}. We use the public implementations of these baseline methods.

We compare our results with five baseline methods on the GTAV dataset. The evaluation scores for the proposed method and compared methods are reported in Table \ref{tab:SSIM}. We can see that DiscoGAN and CoGAN  get the worse scores. CycleGAN and Pix2pix have similar scores and CycleGAN achieves a marginally higher results, while all these methods have worse scores compared to Homo. Our proposed model achieves the best results against all other baselines. The better score indicates that our proposed model generates images which are more realistic and reasonable, and further more similar to the ground truth images in the global structure. 

In order to make intuitive comparisons, we display some representative examples generated by our proposed model and baselines in Fig.~\ref{fig:exper}. In each example, there are eight images, which always appear in the same order; from left to right: frontal view, ground truth, our proposed model, Homo, CycleGAN, DiscoGAN, CoGAN and Pix2pix.  It can be seen that the samples generated by the DiscoGAN and CoGAN are blurry, and the structure of vehicles in the image generated by CoGAN is not correct. CycleGAN generates images with more details, but also presents severe artifacts and incorrect position of object. Pix2pix maintains the basic structure, but severely lacks of details and suffers from the mode collapse problem (e.g. images generated by Pix2pix in Fig.~\ref{fig:exper} (4) and (5)). Moreover, CycleGAN generally remains the homograpy view and suffers from distortion (e.g. images generated by CycleGAN in Fig.~\ref{fig:exper} (3) and (5)). 
Homo could produce reasonable images in the global appearance while it also has a severe distortion, especially for the object (e.g. the vehicle) in a distance. With the global context from frontal view and two consistent constraints, our proposed model produces more realistic images, which keep the global structure consistent with the ground truth and possess richer details and correct color. 
For instance, in the fourth example (Fig.~\ref{fig:exper} (3)), the image generated by our model has more sharper details (e.g. consistent color with the ground truth) and reasonable appearance of vehicle.

\begin{table}
\vspace{-1ex}
\small
\setlength{\tabcolsep}{0.9pt}
\begin{center}
\label{tab:SSIM}
\vspace{-1ex}
\begin{tabular*}{1.0\linewidth}{l|cccccc}
\toprule
Method & DiscoGAN & CycleGAN &CoGAN & Pix2pix &Homo &Ours\\
\midrule
SSIM $\uparrow$ & 0.5342 & 0.5568 & 0.5453 & 0.5486 & 0.5715 & \textbf{0.5961}\\
PSNR $\uparrow$ & 4.8432 & 4.8754 & 4.8602 & 4.8703 & 4.9234 & \textbf{5.0056}\\ 
LPIPS $\downarrow$ & 0.3741 & 0.2902 & 0.3588 & 0.3083 & 0.2709 & \textbf{0.2427}\\
\bottomrule
\end{tabular*}
\caption{The results of our proposed method and baseline methods.}
\end{center}
\vspace{-5ex}
\end{table}

\subsection{Ablation Study}

To verify the contributions of different components in our model, we design several variants to perform an ablation study. The details of these variants are shown as follows:
\begin{itemize}
\item[$\bullet$] {\bf Ours without domain $Z$}: We remove the frontal view $Z$. This variant only contains two GANs representing the left two domains.
\vspace{-1.0ex}
\item[$\bullet$] {\bf Ours without domain $X$}:
We remove the intermediate view $X$. This variant performs cross-view translation directly from the frontal view to bird view.
\vspace{-1.0ex}
\item[$\bullet$] {\bf Ours without dual cycle-consistency}: We remove the dual cycle-consistency loss from the total loss.
\vspace{-0.5ex}
\item[$\bullet$] {\bf Ours without Cross-domain feature consistency}: We remove the cross-domain feature consistency loss from the total loss. 
\vspace{-2ex}
\end{itemize}

\begin{table}
\centering
\small
\setlength{\tabcolsep}{3.5pt}
\begin{center}
\label{tab:ablation}
\begin{tabular*}{1.0\linewidth}{p{1.1cm}|p{1.2cm}<{\centering}|p{1.2cm}<{\centering}|p{1.2cm}<{\centering}|p{1.2cm}<{\centering}|p{0.7cm}<{\centering}}
\toprule
Variant & Ours w/o $Z$ & Ours w/o $X$  &Ours w/o cyc loss  &Ours w/o cfc loss & Ours\\
\midrule
SSIM $\uparrow$ & 0.5726 & 0.5116 & 0.5701 & 0.5842 &\textbf{0.5961}\\
\hline
LPIPS $\downarrow$ & 0.2634 & 0.3225 & 0.2679 & 0.2581 & \textbf{0.2427}\\
\bottomrule
\end{tabular*}
\caption{Results of ablation study. Cyc loss denotes the dual cycle-consistency loss, and cfc is the cross-view feature consistency loss.}
\end{center}
\vspace{-5ex}
\end{table}

Table \ref{tab:ablation} reports the results of ablation study. It can be found that removing the loss function, i.e. dual cycle-consistency or cross-view feature consistency, degrades the results of model, so does removing the domain $X$ or domain $Z$. In particular, removing the domain $X$ greatly worsens the quality of the generated images, which indicates the huge gap between frontal view and bird view and the importance of the intermediate view $X$. We therefore conclude that all three views play a critical role in our model, and two consistency constraints are also important parts of improving quality of generated images.

\begin{figure}[t]
\begin{center}
   \includegraphics[width=0.8\linewidth]{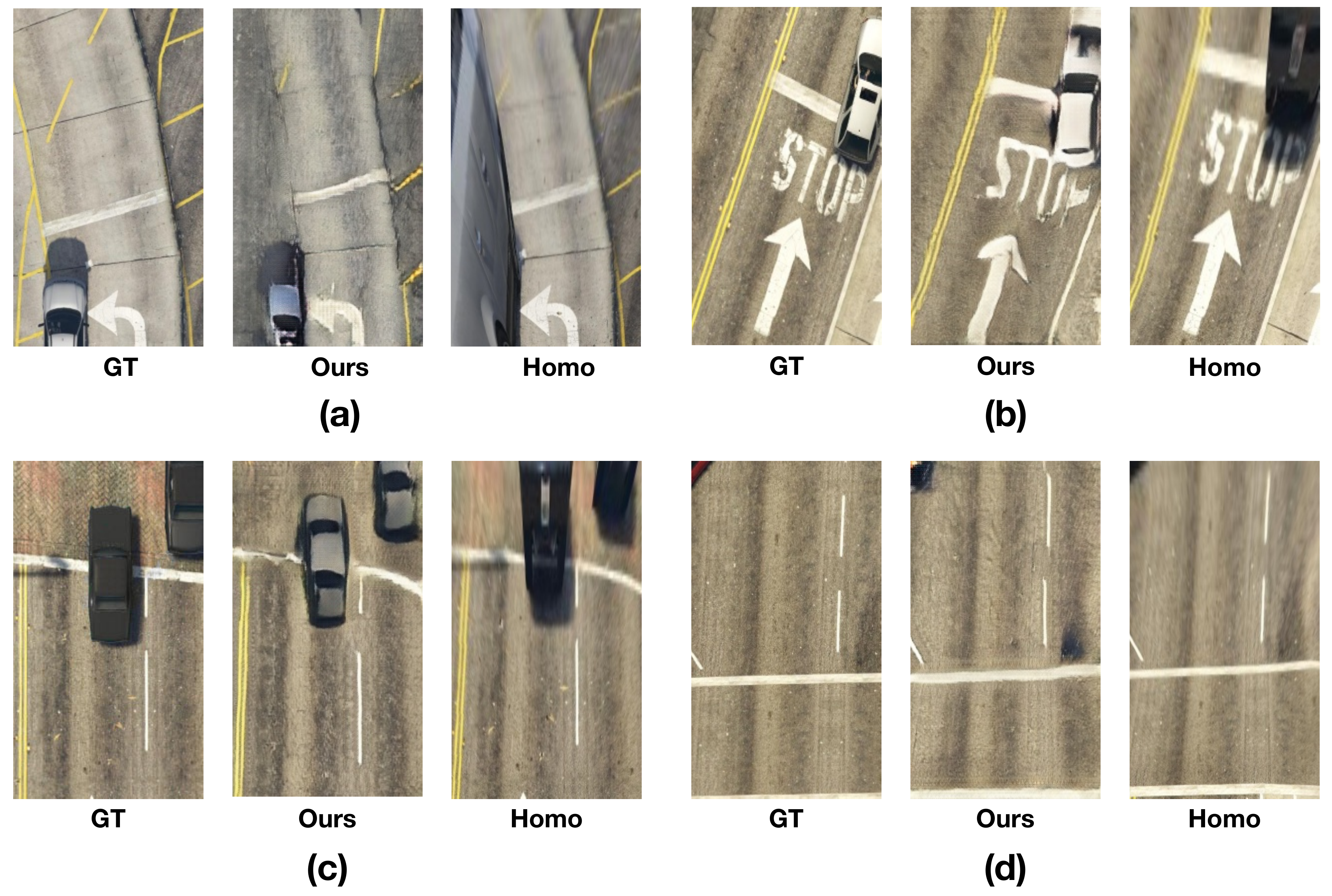}
\end{center}
\vspace{-2ex}
\caption{Some challenging examples. (a): The car is right in front of the camera. (b): The car is on the curved road. (c): There are multiple vehicles. (d): There is no vehicle. In these cases, the homography estimation fails and suffers severe distortions.}
\label{fig:challenge}
\vspace{-2ex}
\end{figure}

\subsection{Discussion}
\noindent \textbf{Could the model work well when the homography view fails?}~~~We can find that the proposed intermediate view bridges the large gap. Hence, could the model work well when the homography view fails or suffers severe distortions?
In order to further verify the capability of our proposed model, we show some challenging samples in Fig.~\ref{fig:challenge}. In Fig.~\ref{fig:challenge}(a) and (b), the vehicles are right in front of the camera and on the curved road, respectively, where the homography view fails and suffers from severe distortions. The proposed model successfully generates the reasonable bird view with the failed homography estimation.

\vspace{1ex}
\noindent \textbf{Does the model synthesize the cars that do not exist?}~~~In the normal GAN model, it might generate diverse outputs even non-exist things given the same input. However, our proposed model would not synthesize a non-exist car because the homography view is introduced as an input source which could provide initial location beliefs on the cars, even for the homography estimation with severe distortions it also supplies the stable beginning position of vehicles. As shown in Fig.~\ref{fig:challenge}(c) and Fig.~\ref{fig:challenge}(d), the proposed model successfully synthesizes a multi-car bird view image and a no-car bird view image under the guidance of the homography view, respectively. It demonstrates that our proposed model is reliable and would not hallucinate vehicles that do not exist.

\subsection{User Study}
Automatic evaluation measures, such as SSIM, could reduce the cost of time and human effort. However, these methods are not always reliable. In \cite{goodfellow2014explaining,fool}, they found that some generated images are not more realistic but score very highly, which indicates that these automatic measures are not fully correlated with human perception. 
Thus, the user study is applied to supplement the evaluation, in which the generated images are evaluated by human observers.

We conducted two types of user study experiments on this dataset for quantitative evaluation. In the first experiment, we showed 110 groups of images. Each group showed one ground truth image, and six images generated by baseline methods and our proposed model based on the same input image. Users were asked to choose an image that is most similar to the ground truth and has the best quality in terms of appearance and detail. In the second experiment, we showed 110 groups of images which were randomly sampled from the whole testing set, and users were asked to choose an image that has the best quality for each group.  A total of 25 users participated in this user study. 

We calculate the percentages of each model whose generated image is selected as the best image. The results of two experiments are shown in Table \ref{tab:user}. It can be observed that our proposed model achieves the best performances in both experiments with 16\% and 7\% higher scores than baselines respectively, which indicates that our model could generate more realistic and reasonable images which better satisfy the human standard.
\begin{table}
\small
\vspace{-1ex}
\setlength{\tabcolsep}{1.0pt}
\begin{center}
\label{tab:user}
\begin{tabular*}{1.0\linewidth}{l|c|c|c|c|c|c}
\toprule
Model & DiscoGAN & CycleGAN  &CoGAN & Pix2pix &Homo &Ours \\
\midrule
Score1 &0.0249 &0.1622   & 0.0134  & 0.0875 &0.2730 & \textbf{0.4390} \\
\midrule
Score2 & 0.0173 & 0.1736  & 0.0251 & 0.0517 & 0.3299 & \textbf{0.4024} \\
\bottomrule
\end{tabular*}
\caption{Results of two types of user study experiments. Score1 is the result for the first experiment and the Score2 is for the second.}
\end{center}
\vspace{-3ex}
\end{table}


\section{Conclusion}
\label{sec:con}

In this paper, we propose a novel cross-view translation model, i.e., BridgeGAN, to address the new problem of bird view synthesis from a single frontal view image. It can provide better perceptual understanding and enable future researchers with multiple views perception attempt.
Specifically, we first introduce an intermediate view to bridge the huge gap between frontal view and bird view. Then the multi-GAN model is proposed to perform the cross-view translation. 
Two constraints are introduced to our proposed model to ensure one-to-one cross-domain translations and consistent cross-view feature representations.
With extensive experiments, our model typically preserves the scene structure, global appearance and details of objects in the bird view.
Quantitative evaluations and user study demonstrate the superiority of the proposed model. Ablation studies verify the importance of each components. More discussions and examples show its reliability even in the challenging cases.

{\small
\bibliographystyle{ieee}
\bibliography{egbib}
}

\end{document}